\documentclass[10pt,twocolumn]{article}

\usepackage{times}             
\usepackage{graphicx}          
\usepackage{amsmath, amssymb}  
\usepackage{caption}           
\usepackage{subcaption}        
\usepackage{geometry}          
\usepackage{authblk}           
\usepackage{fancyhdr}          
\usepackage{abstract}          
\usepackage{titlesec}          
\usepackage{hyperref}          
\titleformat{\section}{\large\bfseries}{\thesection.}{1em}{}
\titleformat{\subsection}{\normalsize\bfseries}{\thesubsection}{1em}{}

\title{\textbf{RT-DETRv2 Explained in 8 Illustrations}}
\author{Ethan Qi Yang CHUA, Jen Hong TAN \\ 
Data Science and Artificial Intelligence Lab \\
Singapore General Hospital}

\date{}

\begin{document}
\twocolumn[
\maketitle
]

\section*{ABSTRACT}
\addcontentsline{toc}{section}{ABSTRACT}
    Object detection architectures are notoriously difficult to understand, often more so than large language models. While RT-DETRv2 \cite{lv2024rtdetrv2improvedbaselinebagoffreebies} represents an important advance in real-time detection, most existing diagrams do little to clarify how its components actually work and fit together. In this article, we explain the architecture of RT-DETRv2 through a series of eight carefully designed illustrations, moving from the overall pipeline down to critical components such as the encoder, decoder, and multi-scale deformable attention. Our goal is to make the existing one genuinely understandable. By visualizing the flow of tensors and unpacking the logic behind each module, we hope to provide researchers and practitioners with a clearer mental model of how RT-DETRv2 works under the hood.

\section{INTRODUCTION}
    Object detection plays a critical role in computer vision, with models such as YOLO \cite{redmon2016lookonceunifiedrealtime} achieving efficiency by framing detection as a single-pass regression task over dense anchor boxes, followed by heuristic post-processing like non-maximum suppression (NMS). While fast, these approaches often struggled with spatial context and multi-scale objects.
    
    The introduction of DETR (DEtection TRansformer) \cite{carion2020endtoendobjectdetectiontransformers} marked a shift: object detection was reformulated as a set-prediction problem, where global attention and bipartite matching directly produce object queries. Although elegant, DETR converges slowly and has difficulty with small objects due to its reliance on global attention.
    
    RT-DETRv2 \cite{zhao2024detrsbeatyolosrealtime} was designed to overcome these limitations. It incorporates deformable attention, hybrid encoding, denoising, query-based decoding, and other refinements to achieve real-time performance without anchor heuristics. Despite these advances, the architecture of RT-DETRv2 remains complex, and existing diagrams in the literature rarely make the tensor flow or design choices easy to grasp.
    
    In this article, we aim to demystify RT-DETRv2. Using eight carefully designed illustrations, we progressively break down the model—starting from the overall pipeline, through the encoder and decoder, and down to intricate components like multi-scale deformable attention. Our goal is not to introduce new methods, but to make an important model accessible by visualizing how its parts truly fit together.
    
\section{METHOD}
\begin{figure}[h]
    \centering
    \includegraphics[width=\linewidth]{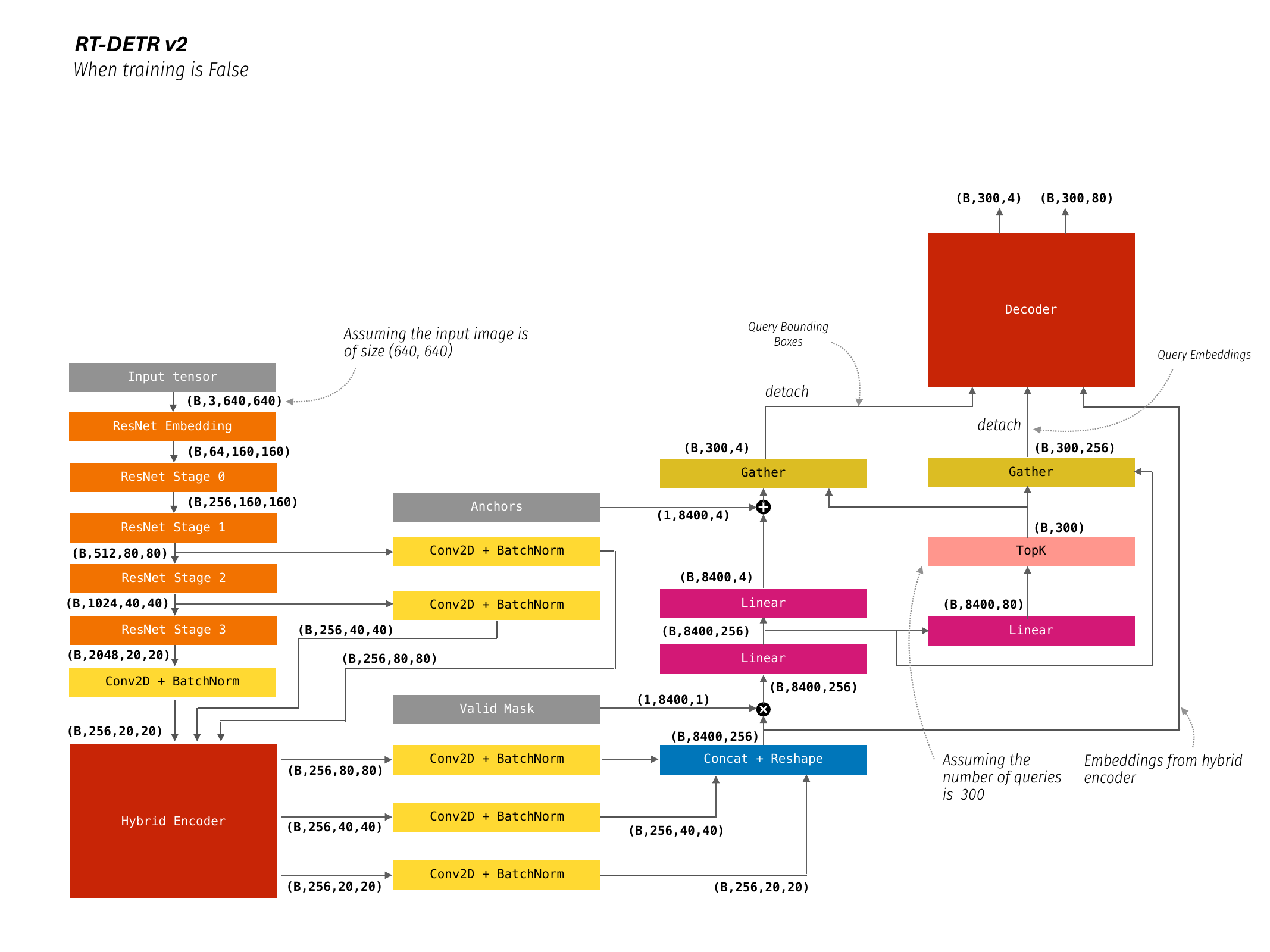}
    \caption{The structure of an RT-DETRv2 model with denoising task off.}
    \label{fig:rtdetrv2off}
\end{figure}

\begin{figure}[h]
    \centering
    \includegraphics[width=\linewidth]{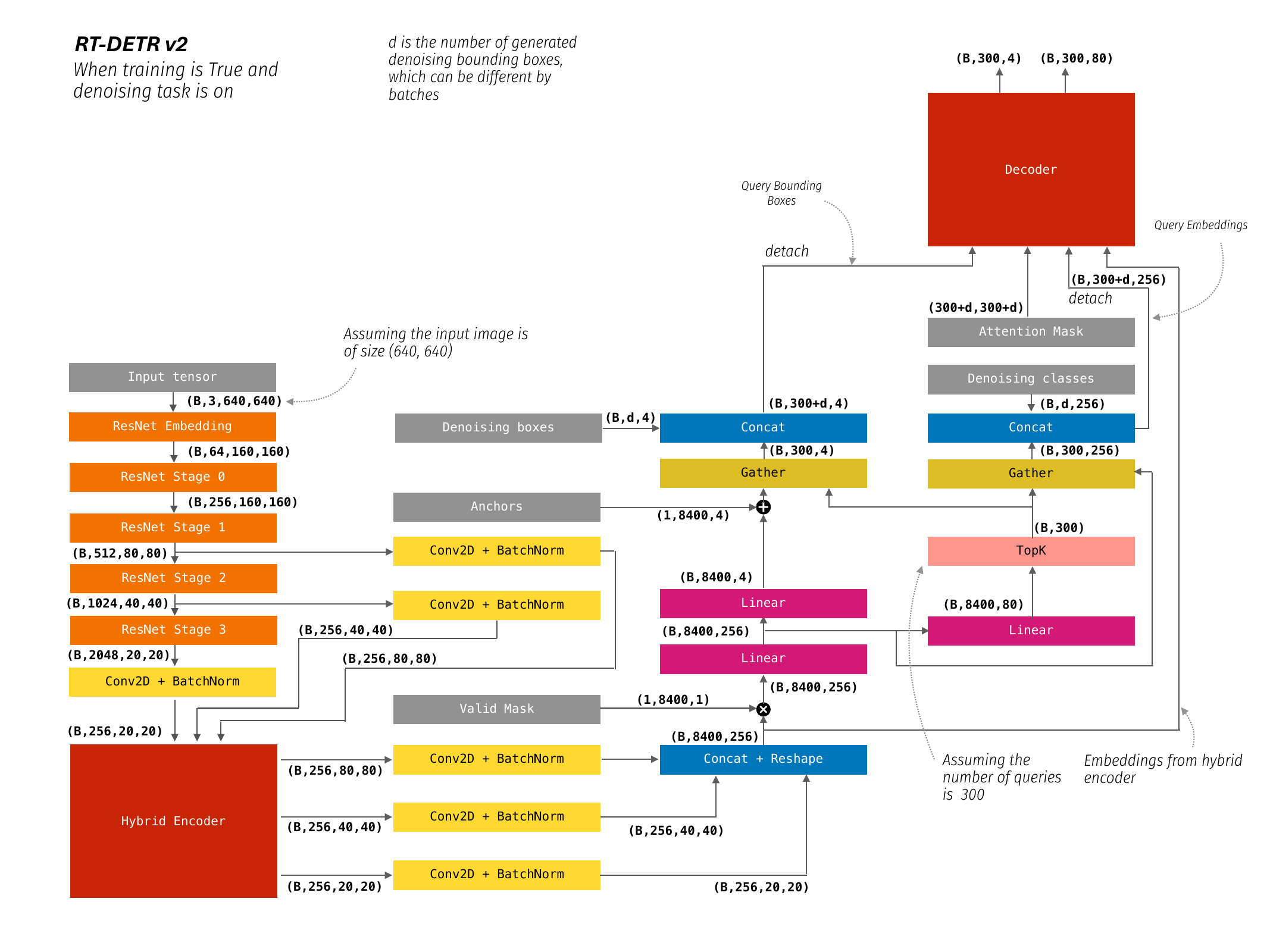}
    \caption{The structure of an RT-DETRv2 model with denoising task on.}
    \label{fig:rtdetrv2on}
\end{figure}

\begin{figure}[h]
    \centering
    \includegraphics[width=\linewidth]{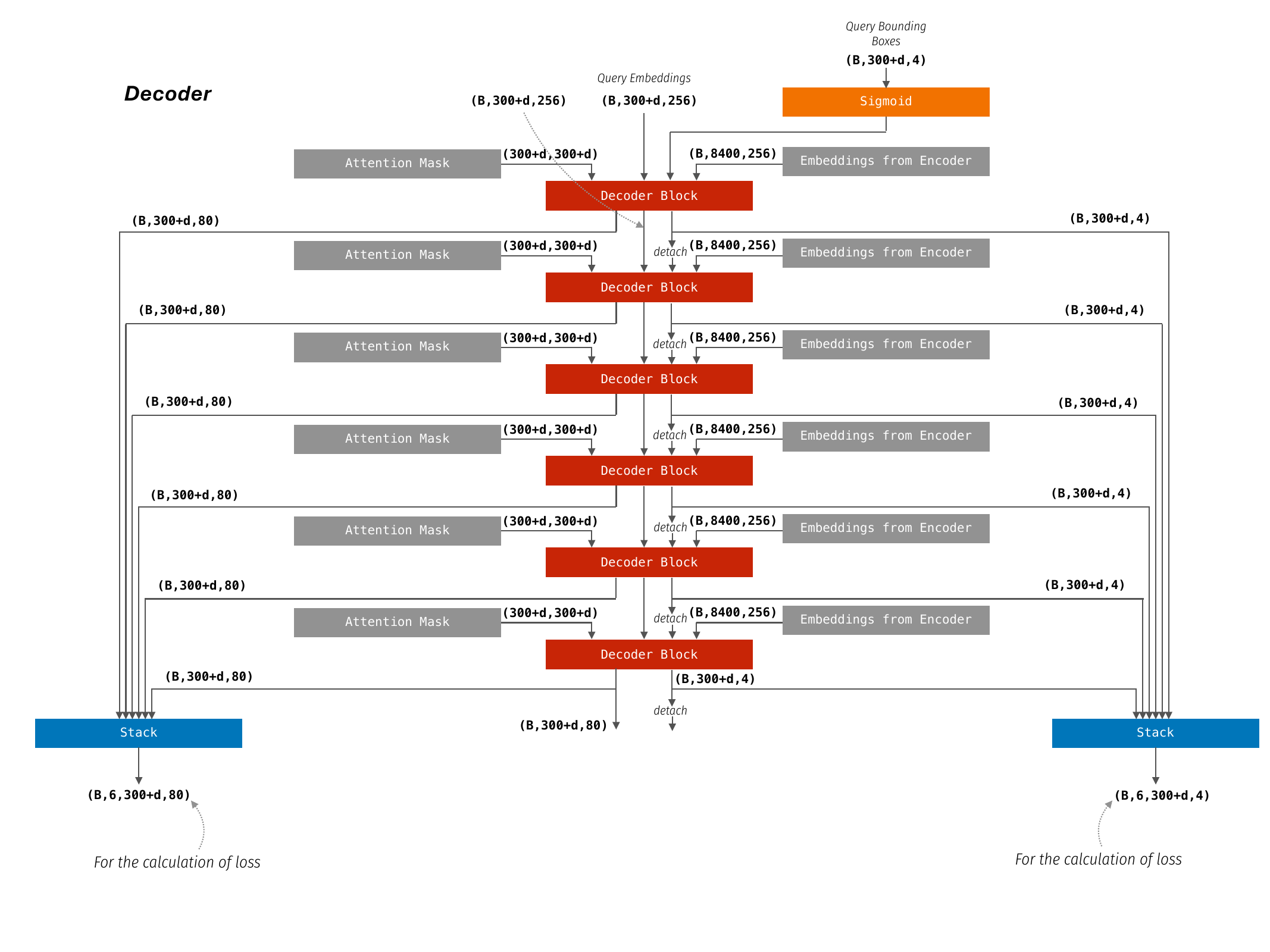}
    \caption{The structure of the decoder.}
    \label{fig:decoder}
\end{figure}

\begin{figure}[h]
    \centering
    \includegraphics[width=\linewidth]{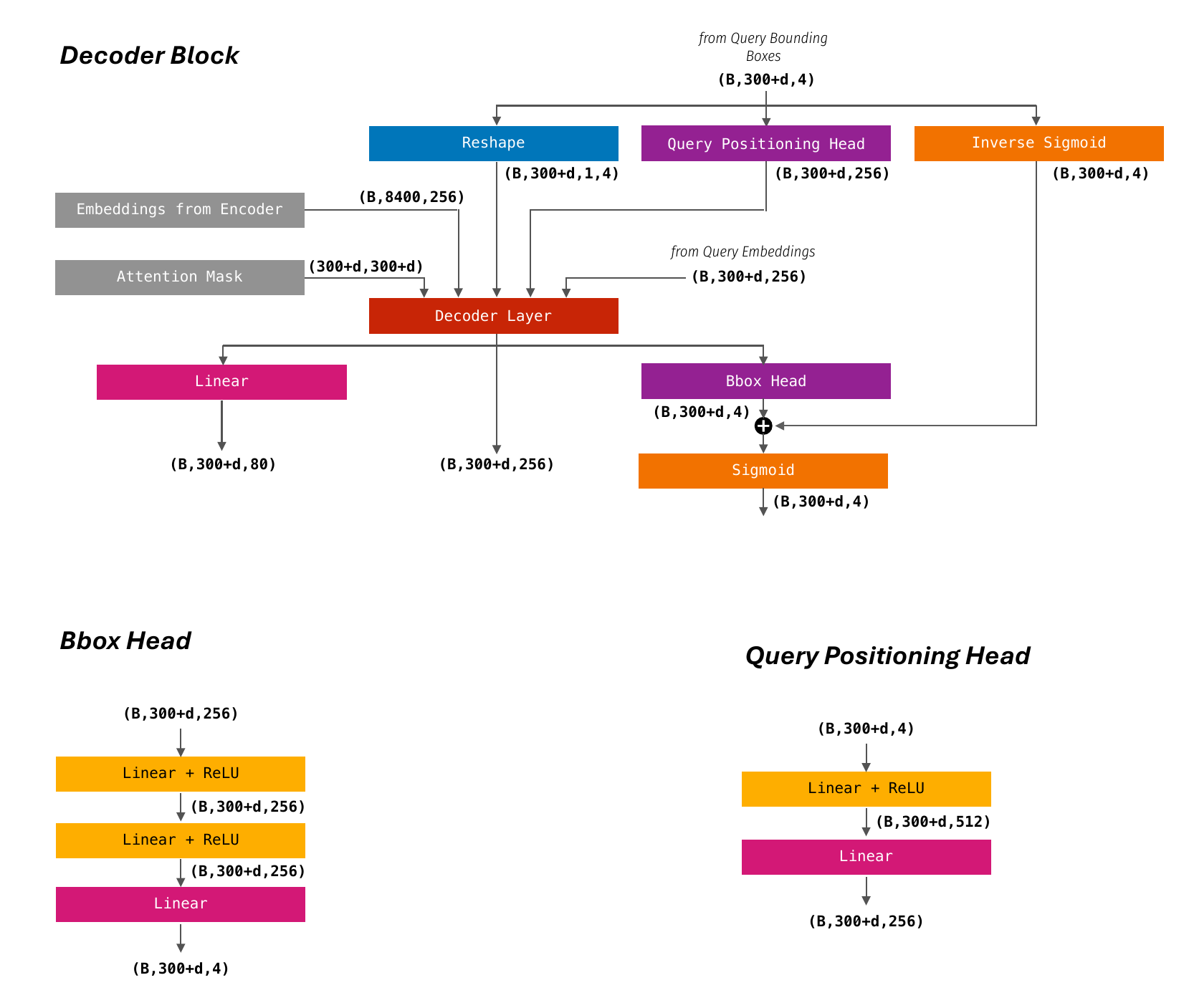}
    \caption{Above is the structure of the decoder block. Bottom left is the bbox head structure. Bottom right is the query positioning head structure.}
    \label{fig:decoderblock}
\end{figure}

\begin{figure}[h]
    \centering
    \includegraphics[width=\linewidth]{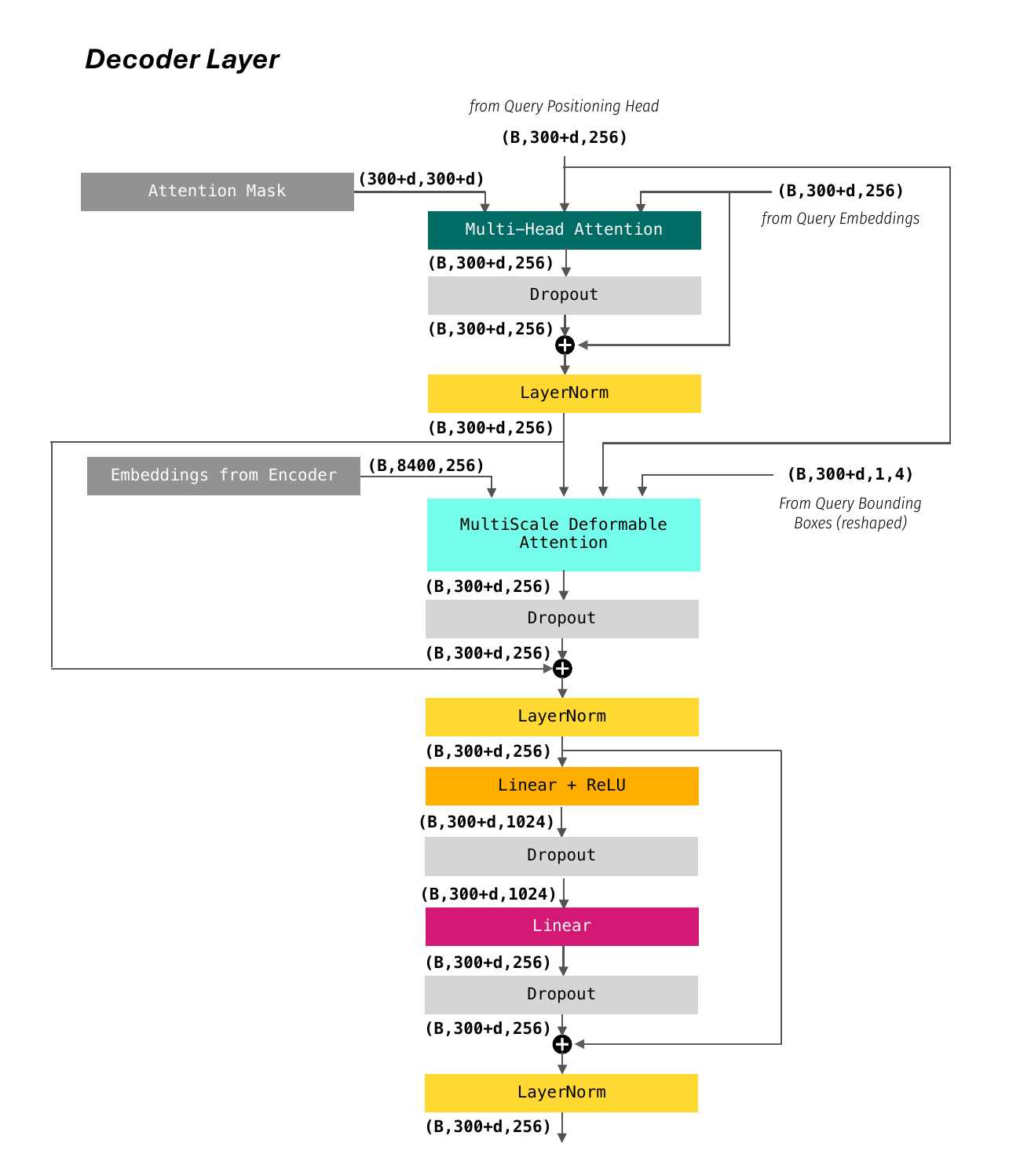}
    \caption{The structure of the decoder layer.}
    \label{fig:decoderlayer}
\end{figure}

    The RT-DETRv2 architecture can be divided into five main components: the CNN backbone, hybrid encoder, denoising task, query selection, and decoder. In this article, ResNet50 \cite{he2015deepresiduallearningimage} is the CNN backbone, and we follow the default configuration described in the original paper/code in Transformers library, using input images of size $(640, 640)$, $300$ object queries, and $100$ denoising queries (though the actual number of denoising bounding boxes, denoted by $d$ from Figure to \ref{fig:rtdetrv2on} to \ref{fig:multiscaledeformableattention}, varies across batches). We use $B$ to denote batch size. Figure \ref{fig:rtdetrv2off} illustrates the overall tensor flow in RT-DETRv2 during inference, where the denoising task is disabled.

\subsection{CNN Backbone}
    The input image first passes through a ResNet-based embedding layer followed by several ResNet stages \cite{he2015deepresiduallearningimage}, producing three multi-scale feature maps of sizes $(B, 256, 80, 80)$, $(B, 256, 40, 40)$, and $(B, 256, 20, 20)$. These feature maps capture information at different spatial resolutions: the $80 \times 80$ map preserves fine-grained spatial detail but carries relatively shallow semantic context, whereas the $20 \times 20$ map provides coarse spatial resolution with richer semantic representation. The intermediate $40 \times 40$ map strikes a balance between the two, supporting effective multi-scale feature integration in later stages of the model.

\subsection{Hybrid Encoder}
\begin{figure}[h]
    \centering
    \includegraphics[width=\linewidth]{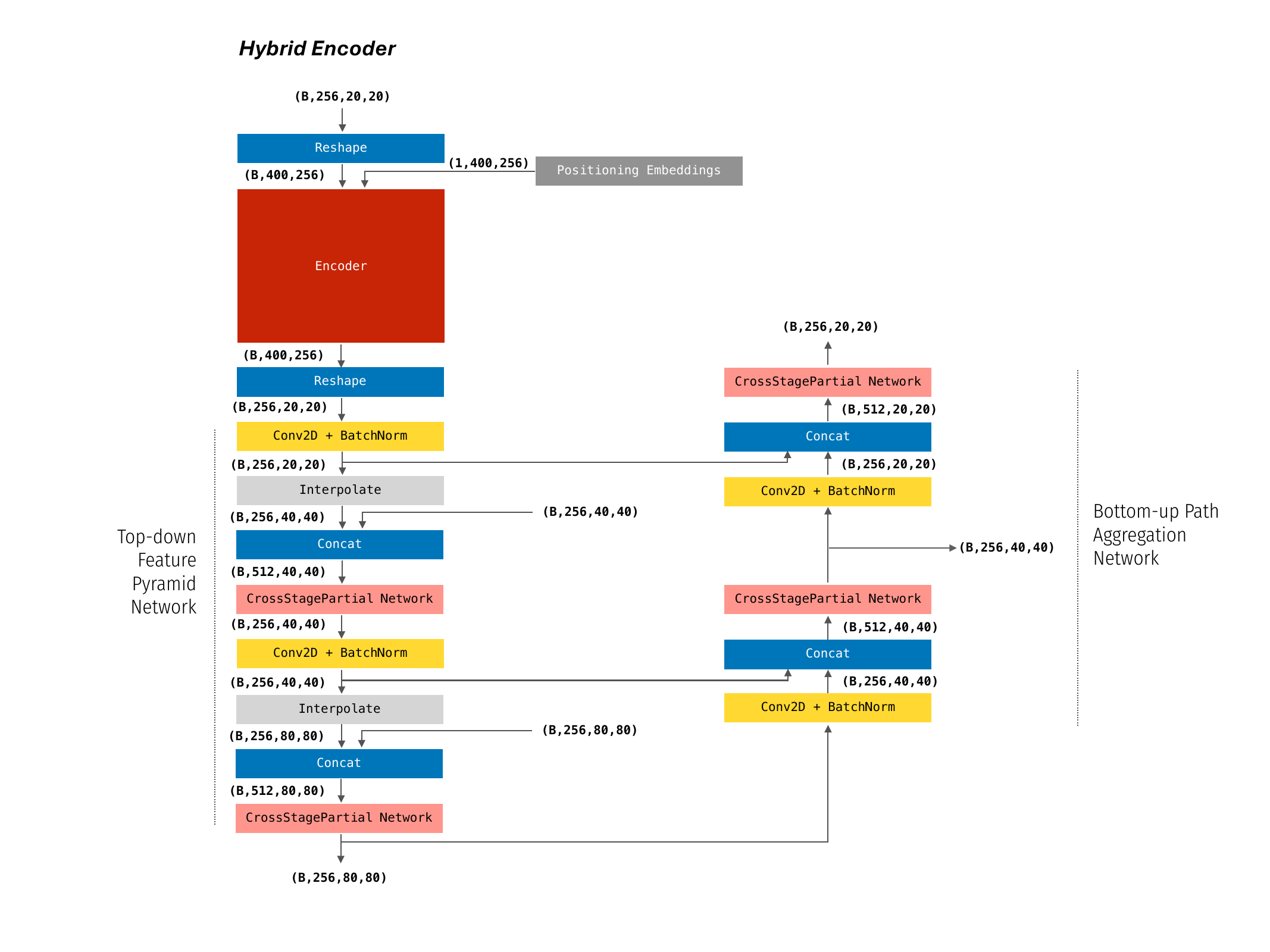}
    \caption{The structure of the hybrid encoder. The hybrid encoder is split into two distinct components: the encoder which performs normal self-attention, and the fusion pathways as seen by the Top-Down Feature Pyramid and Bottom-Up Path Aggregation Networks.}
    \label{fig:hybridencoder}
\end{figure}

    The hybrid encoder \cite{yang2021hybridencoderefficientprecise} consists of an encoder and a series of fusion pathways (Figure \ref{fig:hybridencoder}). The encoder performs self-attention \cite{vaswani2023attentionneed} only on the feature map with lowest resolution and highest semantic richness to understand relationships and identify object features globally. The fusion pathways perform top-down and bottom-up feature aggregation \cite{lin2017featurepyramidnetworksobject} through the use of upsampling, downsampling, concatenation, and Cross Stage Partial Networks \cite{wang2019cspnetnewbackboneenhance} to ensure that high resolution features also contain deep semantic meaning.

    The Top-Down Feature Pyramid and Bottom-Up Aggregation Network are fusion pathways which aim to produce feature maps which are high in semantic richness and spatial resolution.

\begin{figure}[h]
    \centering
    \includegraphics[width=\linewidth]{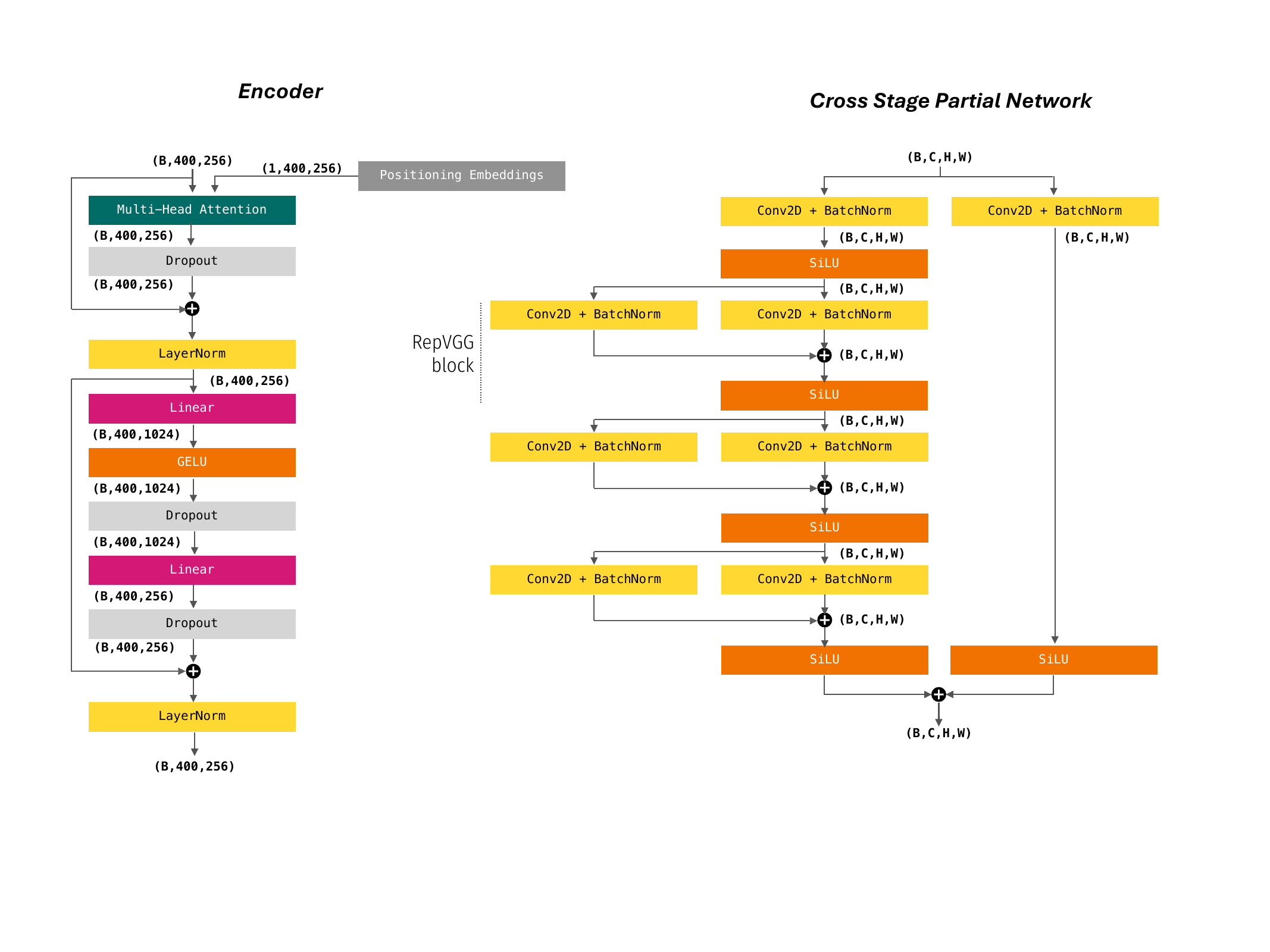}
    \caption{Left is the structure of the encoder, performing rather straightforward self-attention on the input tensor. Right is structure of the Cross Stage Partial Network which utilises RepVGG blocks \cite{ding2021repvggmakingvggstyleconvnets} to allow for faster convergence and fewer vanishing gradients in deep neural networks.}
    \label{fig:cspnet}
\end{figure}

    For the Top-Down Feature Pyramid, the $(B, 256, 20, 20)$ tensor is returned by the encoder and upsampled to size $(40, 40)$. It is concatenated with the $(B, 256, 40, 40)$ tensor from ResNet Stage 2 (Figure \ref{fig:rtdetrv2off}) resulting in $512$ feature maps $(256 + 256)$. This is passed through a Cross Stage Partial (CSP) Network (Figure \ref{fig:cspnet}) to output a tensor of shape $(B, 256, 40, 40)$. The process repeats, where the tensor is upsampled and concatenated with the $(B, 256, 80, 80)$ feature maps from ResNet Stage 3 to result again in $512$ feature maps. It is passed through another CSP Net to output a tensor of shape $(B, 256, 80, 80)$.

    The Bottom-Up Path Aggregation Network operates similar to the Top-Down Feature Pyramid, just that it downsamples the tensors instead. The $(B, 256, 80, 80)$ tensor is then downsampled to size $(40, 40)$ and concatenated with the earlier $(B, 256, 40, 40)$ CSP Net output to produce a tensor with $512$  maps. The tensor goes through the same process to output shape $(B, 256, 40, 40)$. These steps are repeated to output another tensor of shape $(B, 256, 20, 20)$. The resultant tensors of size $(80, 80)$, $(40, 40)$ and $(20, 20)$ each have $256$ channels. These fused representations are the semantically and spatially rich outputs of the hybrid encoder for the rest of the model.

\subsection{Query Selection}
The purpose of query selection is to reduce the dense set of candidate features into a smaller set of informative queries for the decoder. Once the $(B, 256, 80, 80)$, $(B, 256, 40, 40)$, and $(B, 256, 20, 20)$ feature maps have been returned by the hybrid encoder, they are concatenated and reshaped to a $(B, 8400, 256)$ tensor where $8400$ represent the spatial locations across all scales and $256$ represent the number of channels for each spatial location. A valid mask of shape $(1, 8400, 1)$ is then applied through element-wise multiplication to the tensor in order to omit any invalid spatial locations. The tensor is then passed through a linear layer to map feature embeddings to classification logits, resulting in shape $(B, 8400, 256)$.

Afterwards, the tensor is passed through a classification branch and a localisation branch. The classification branch passes the tensor through a linear layer resulting in $(B, 8400, 80)$ whereby the prediction ranges between 80 object classes. The $(B, 8400, 80)$ tensor is then passed through a TopK algorithm to gather the top 300 queries which likely contain important information to be input into the decoder.

The localisation branch is passed through a linear layer resulting in $(B, 8400, 4)$ where the last dimension represents the coordinates $(x_1, y_1, x_2, y_2)$. Anchors are added to the tensor at the localisation branch before the TopK queries are filtered out. Query selection ensures that only the most relevant queries are forwarded to the decoder, reducing redundancy while preserving spatial diversity.

\subsection{Denoising Task}
When the denoising task \cite{li2022dndetracceleratedetrtraining} is enabled, the model takes certain ground truth images and labels and deliberately alters them slightly to create denoising queries. The denoising queries remove the need for matching since they directly correlate to the ground truth bounding boxes and labels. The denoising queries will have their own loss computed separately from the main queries.

\subsection{Decoder}
The decoder consists of six decoder blocks which takes the query embeddings, query bounding boxes, encoder embeddings, and attention masks as inputs which result in class and bounding box predictions as the output.

The decoder also has two stacks which compute the loss for each decoder block individually. The 'detach' label between each decoder block indicates that for bounding box prediction, there is no backpropagation through the decoder blocks, but rather gradient descent is directly computed from the loss stack for each individual block.

\subsubsection{Decoder Block}

Zooming in to the decoder block, the query bounding boxes are reshaped to $(B, 300+d, 1, 4)$, where the additional dimension of $1$ ensures compatibility with the computations required by multi-scale deformable attention. The inverse sigmoid is used to move bounding box coordinates from a bounded space of $[0,1]$ to an unbounded space where the decoder can adjust the coordinates freely before passing it through the sigmoid function again to keep the prediction bounded.

The bbox head has the tensor pass through two linear + ReLU layers and a final linear layer to output a tensor of shape $(B, 300+d, 4)$ with $4$ coordinates per query. Since this is added element-wise with the output tensor of the inverse sigmoid function, the bbox head outputs positional offsets rather than the raw image coordinates, leading to a more precise result after the resultant tensor is passed through the sigmoid function.

The query positioning head has the tensor pass through a linear + ReLU layer to expand the tensor to a richer $512$ dimensional representation before reducing it through a linear layer to the decoder's $256$ dimensions.

\subsubsection{Decoder Layer}
The decoder layer is central to the decoder block as it processes the encoder embeddings, query embeddings, attention mask, query bounding boxes, and query positional embeddings.

The decoder layer takes the tensors from the query positioning head $(B, 300+d, 256)$, query embeddings $(B, 300+d, 256)$ and the attention mask $(300+d, 300+d)$, and put its through self-attention to refine the embeddings by exchanging contextual information, producing a tensor of shape $(B, 300+d, 256)$. The output is then passed through multi-scale deformable attention \cite{xia2022visiontransformerdeformableattention} where each query predicts a set of spatial reference points for the $(B, 8400, 256)$ encoder feature map. The model then performs multi-scale deformable attention by attending only to the selected reference points, this selective attention allows for more precise detection since there is greater focus on areas of interest in the picture.

The output from the multi-scale deformable attention is added element-wise with the output from the query self-attention as a residual connection and passed through a feed-forward network to predict a tensor of (B, number of queries, embedding dimensions).

\subsubsection{Multi-Scale Deformable Attention}

\begin{figure}[h]
    \centering
    \includegraphics[width=\linewidth]{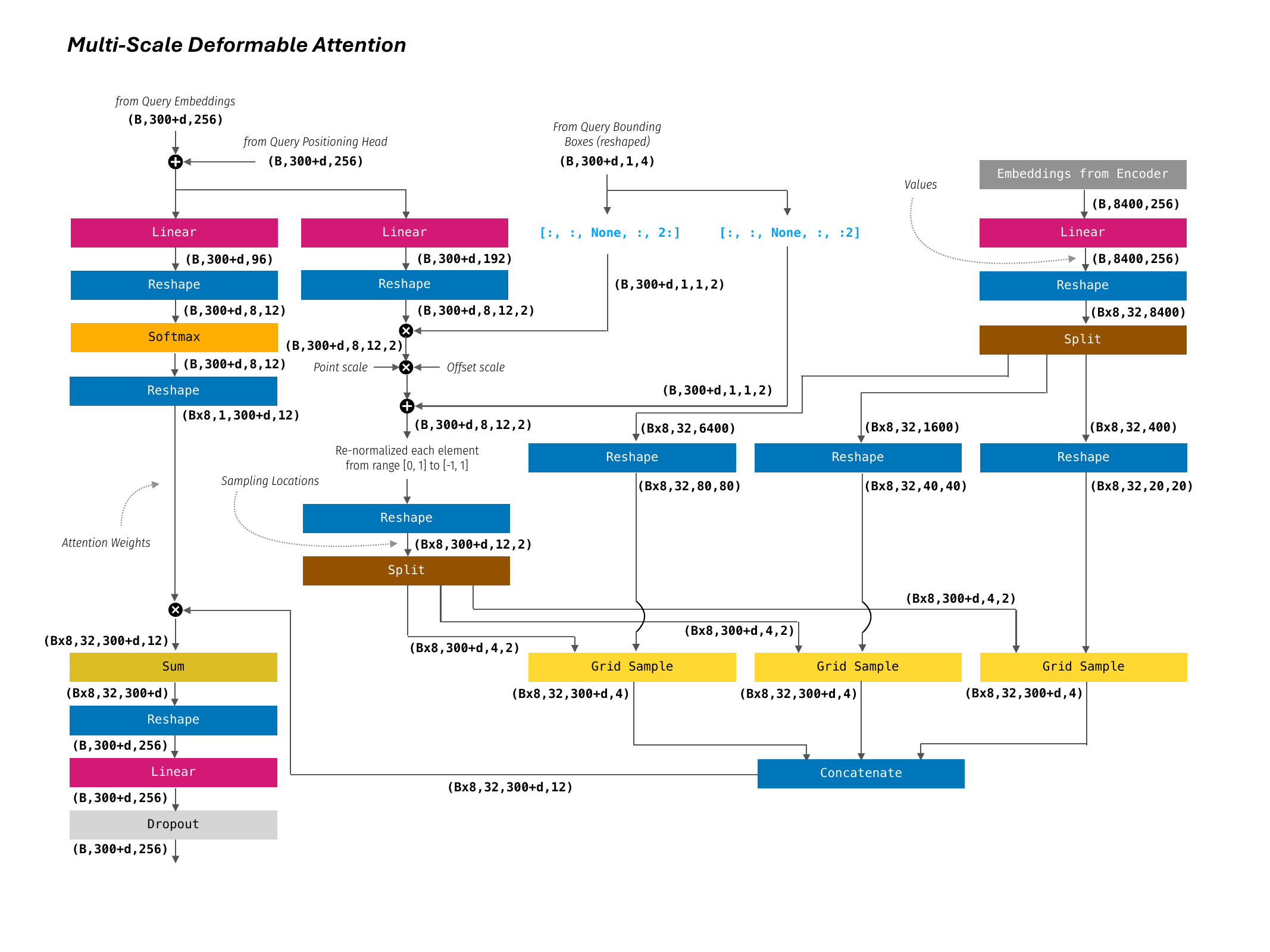}
    \caption{The structure for Multi-Scale Deformable Attention.}
    \label{fig:multiscaledeformableattention}
\end{figure}

In normal self-attention, each query computes similarity scores with all keys to determine attention weights. This global mechanism becomes computationally expensive for tasks in real-time object detection. In multi-scale deformable attention, the attention weights are instead learned through back-propagation. 

In the default architecture with a ResNet-50 backbone, for each query (there are $300+d$ of them when the denoising task is enabled), the total number of sampling locations to be attended to is $96$: for each attention head, there are 4 sampling points, and there are 8 attention heads per scale, with a total of 3 scales. In Figure \ref{fig:multiscaledeformableattention}, this is reflected in the linear layer that transforms the tensor from shape $(B,300+d,256)$ to $(B,300+d,96)$.

On the other hand, the offsets for each sampling location ($96$ in total) are learned and produced by another linear layer. In Figure \ref{fig:multiscaledeformableattention}, this corresponds to the linear layer that transforms the tensor from shape $(B,300+d,256)$ to $(B,300+d,192)$, since each sampling location is represented by two values, $x$ and $y$. The width and height of the query bounding boxes (in Figure \ref{fig:multiscaledeformableattention}, denoted by \texttt{[:, :, None, :, 2:]}) are involved in the process of generating these offsets. Each offset is then added to its corresponding query bounding box position, yielding the 96 sampling locations for each query bounding box. These sampling locations, now collated into a tensor of shape $(B,300+d,8,12,2)$, are re-normalized from the range $[0,1]$ to $[-1,1]$ a step required for the subsequent grid sampling computation.

On the encoder side, the output tensor of shape $(B,8400,256)$ is reshaped and split into three tensors—each corresponding to one of the three scales. Each scale-specific feature map contains 32 channels per attention head $(256/8)$. The sampling locations tensor is similarly split by scale, and grid sampling is applied to extract the values from the appropriate locations in each feature map. The sampled features across all scales are then concatenated along the point dimension, weighted by the attention weights, summed across sampling points, and reshaped to produce the output tensor of shape $(B,300+d,256)$. A final linear projection and dropout layer complete the operation.

\section{CONCLUSION}
RT-DETRv2 represents a powerful yet intricate architecture for real-time object detection. By unpacking the model step by step, we hope this article provides researchers and practitioners with a clearer understanding of how RT-DETRv2 works under the hood, and serves as a visual reference for future explorations into transformer-based object detection.

\renewcommand{\refname}{4. REFERENCES}
\bibliographystyle{unsrt}
\bibliography{main}

\end{document}